\documentclass[sigconf]{acmart}
\AtBeginDocument{%
  }

\usepackage{amsmath}
\usepackage{mathtools}
\usepackage{amsthm}
\usepackage{hyperref}
\usepackage{url}

\usepackage{graphicx}
\usepackage{subcaption}
\usepackage{paralist}
\usepackage{tabularx}
\usepackage{multirow}
\usepackage{booktabs}
\usepackage{xcolor}
\usepackage{colortbl}
\usepackage{wrapfig}
\usepackage{thm-restate}

\newcommand{\ie}{\emph{i.e., }}
\newcommand{\eg}{\emph{e.g., }}

\theoremstyle{plain}

\theoremstyle{definition}

\theoremstyle{remark}

\usepackage[textsize=tiny]{todonotes}

\copyrightyear{2026}
\acmYear{2026}
\setcopyright{cc}
\setcctype{by}
\acmConference[MM '26]{Proceedings of the 34th ACM International Conference on Multimedia}{November 10--14, 2026}{Rio de Janeiro, Brazil}
\acmBooktitle{Proceedings of the 34th ACM International Conference on Multimedia (MM '26), November 10--14, 2026, Rio de Janeiro, Brazil}
\acmDOI{10.1145/3767308.3836113}
\acmISBN{979-8-4007-2213-4/2026/11}

\begin{document}

\title{PEA-DPO: Perception-Enhanced Alignment Direct Preference Optimization for MLLMs Alignment}

%%
%% Camera-ready author block.
%% Replace the placeholder below with the final author list in exactly the
%% same order as the accepted submission/eRights form. Include each author's
%% institution, city, country, email, and ORCID where available.
\author{Jiawei Feng}
\email{jwf3ng@mail.ustc.edu.cn}
\affiliation{%
  \institution{University of Science and Technology of China}
  \city{Hefei}
  \country{China}}

\author{Jiancan Wu}
\authornote{Corresponding Author.}
\email{wujcan@gmail.cn}
\affiliation{%
  \institution{University of Science and Technology of China}
  \city{Hefei}
  \country{China}}

\author{Xingyu Zhu}
\authornotemark[1]
\email{xingyu.zhu@nus.edu.sg}
\affiliation{%
  \institution{National University of Singapore}
  \city{Kent Ridge}
  \country{Singapore}}

\author{Junkang Wu}
\email{jkwu0909@mail.ustc.edu.cm}
\affiliation{%
  \institution{University of Science and Technology of China}
  \city{Hefei}
  \country{China}}

\author{Xiang Wang}
\email{xiangwang@ustc.edu.cn}
\affiliation{%
  \institution{University of Science and Technology of China}
  \city{Hefei}
  \country{China}}

\author{Xiangnan He}
\email{hexn@ustc.edu.cn}
\affiliation{%
  \institution{University of Science and Technology of China}
  \city{Hefei}
  \country{China}}

\renewcommand{\shortauthors}{Feng et al.}


\begin{abstract}
    Direct Preference Optimization (DPO) has emerged as an effective approach for aligning large language models (LLMs) with human preferences. However, its adaptation to multimodal settings remains unexplored. Through representational analysis, we identify a key limitation in multimodal preference optimization, which we term \textbf{visual insensitivity}: models often fail to distinguish between images and those with critical visual context removed. Our theoretical analysis further uncovers two manifestations of this problem, namely \textbf{Across-Image Insensitivity} and \textbf{Within-Image Insensitivity}. To address these challenges, we propose Perception-Enhanced Alignment DPO (PEA-DPO), a framework for multimodal LLMs alignment, which explicitly leverages visual preference signals to overcome visual insensitivity. We further provide a theoretical analysis demonstrating that PEA-DPO provably mitigates both failure modes. Empirical results demonstrate that PEA-DPO enhances sensitivity to visual context while preserving the language modeling capacity of the base model. Evaluations across three hallucination benchmarks using MLLMs of varying scales show that PEA-DPO effectively mitigates visual insensitivity, achieves stronger multimodal alignment, and substantially reduces hallucinations.
\end{abstract}

%%
%% ACM Computing Classification System concept.
%% Add the matching CCSXML generated at https://dl.acm.org/ccs if TAPS or
%% the production instructions explicitly require the machine-readable block.
% \ccsdesc[500]{Computing methodologies~Artificial Intelligence}



% \keywords{Do, Not, Use, This, Code, Put, the, Correct, Terms, for,


\maketitle

\section{Introduction}

\begin{figure}[htbp]
    \centering
    % 子图 1
    \begin{subfigure}{0.32\linewidth} % 每个子图宽度约占一行的 1/3
        \centering
        \includegraphics[width=\linewidth]{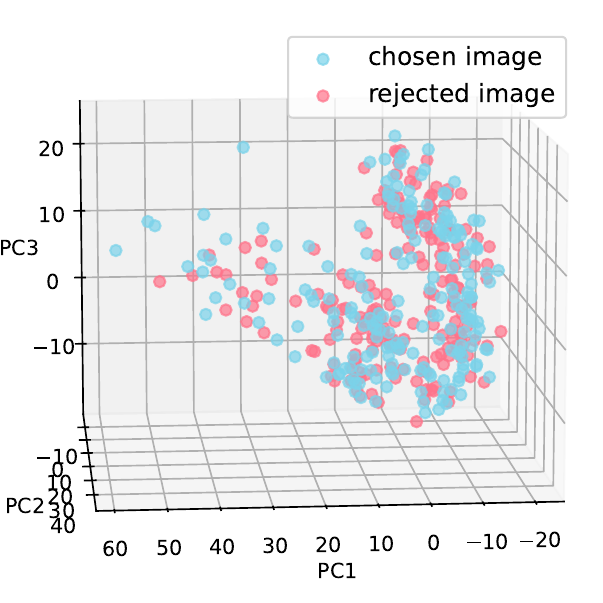}
        \vspace{-2em}
        \caption{LLaVA}
        \label{fig:LLaVA_pca}
    \end{subfigure}
    \hfill
    % 子图 2
    \begin{subfigure}{0.32\linewidth}
        \centering
        \includegraphics[width=\linewidth]{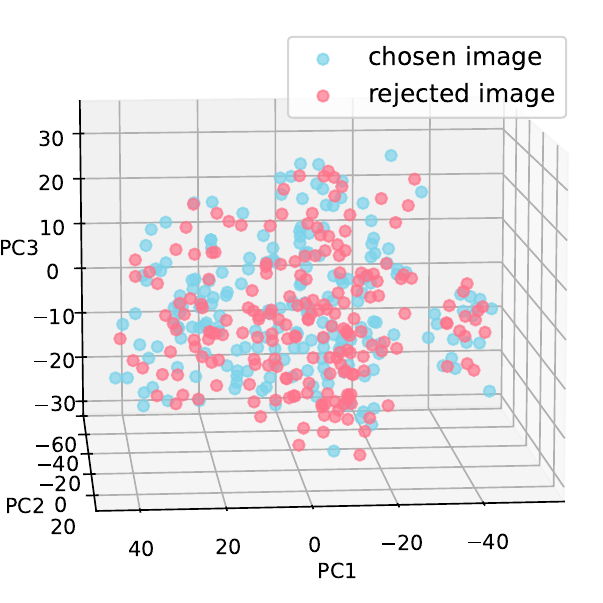}
        \vspace{-2em}
        \caption{LLaVA+DPO}
        \label{fig:subfig2}
    \end{subfigure}
    \hfill
    % 子图 3
    \begin{subfigure}{0.32\linewidth}
        \centering
        \includegraphics[width=\linewidth]{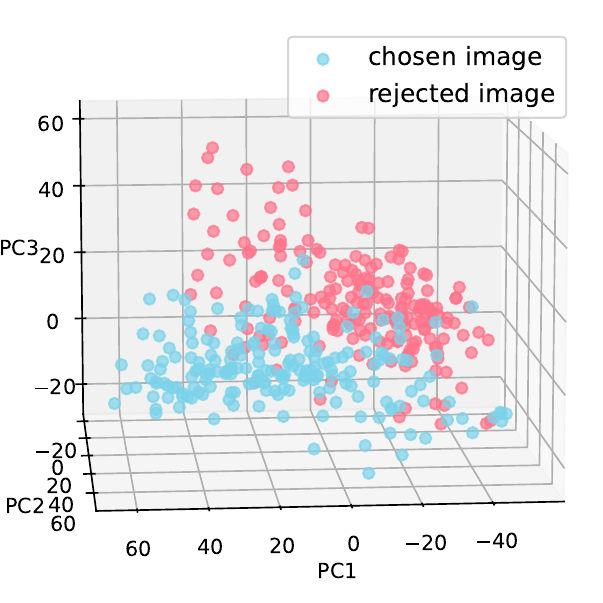}
        \vspace{-2em}
        \caption{LLaVA+PEA-DPO}
        \label{fig:LLaVA+PEA-DPO}
    \end{subfigure}
    \vspace{-1em}
    \caption{Comparison of representation distributions for different models. Representations are constructed from 200 samples (original images and images with removed key visual context), using the embedding of the last token from the LLM to represent image semantics.}
    \Description{Three two-dimensional representation plots compare LLaVA, LLaVA with DPO, and LLaVA with PEA-DPO. Original and context-reduced images overlap for the first two models and are more clearly separated by PEA-DPO.}
    \label{fig:representation_analysis}
\end{figure}

Aligning multimodal large language models (MLLMs)~\citep{zhu2026guardalign, zhu2026look, li2025resbenchbenchmarkingrobustnessmultimodal} with human values is crucial for building reliable AI systems that can understand and reason about visual-textual content while generating helpful responses ~\citep{yin2024survey, bai2025qwen2, liu2024improved}.
The prevailing approaches in this field follow the advances established by text-only language model alignment ~\citep{pi2024strengthening,sarkar2024mitigating, ZhuFWZWYH26, zhu2026robustifying}, applying Direct Preference Optimization (DPO) ~\citep{rafailov2023direct} and its variants ~\citep{fu2025chip, meng2024simpo, yang2025mitigating, lu2025damo} to multimodal scenarios. They typically first construct multimodal preference datasets by pairing images with corresponding preferred and dispreferred textual responses, then apply standard DPO loss to align model outputs with human judgments ~\citep{li2023silkie,xiao2024detecting,zhao2023beyond,zhou2024wpo,yu2024rlhf,deng2024enhancing, zhu2025dynamic, zhu2024enhancing}.

\begin{figure}[htbp]
    \centering
    % ======= subfig 1 =======
    \begin{subfigure}{\linewidth}
        \centering
        \includegraphics[width=\linewidth]{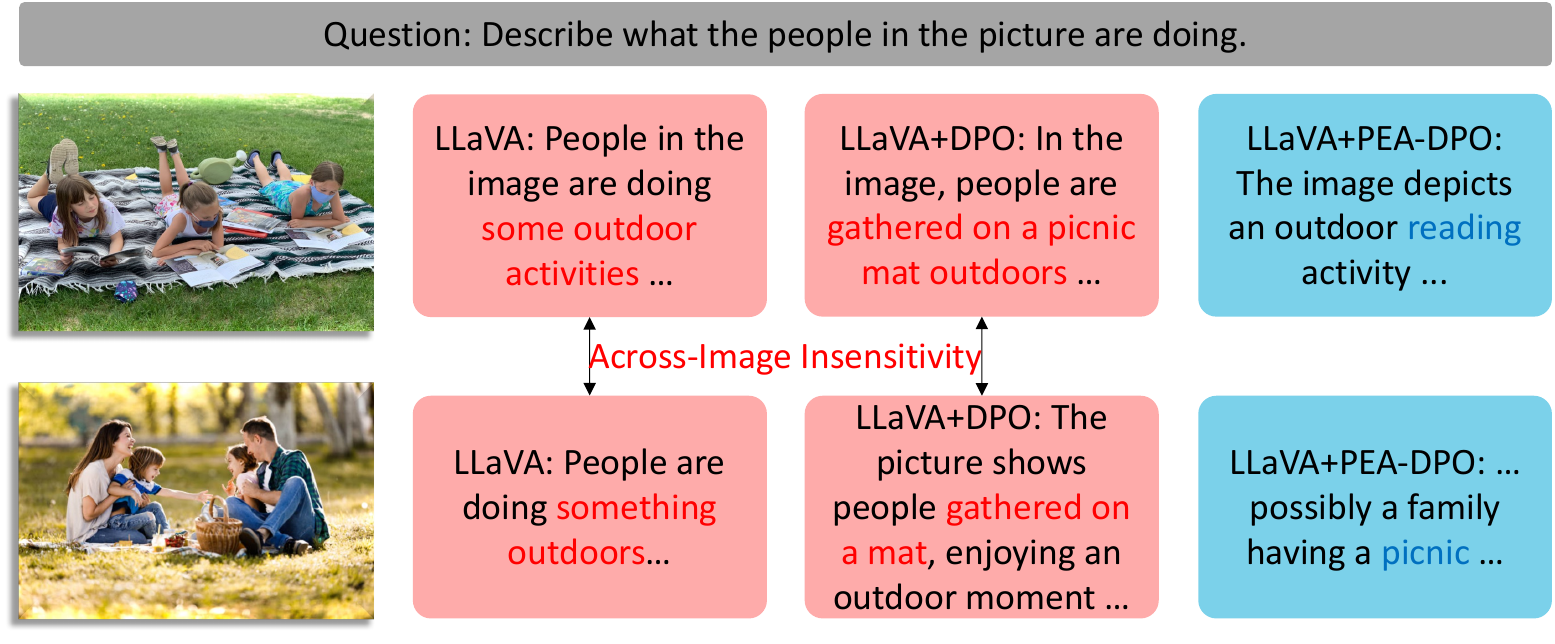}
        \vspace{-1em}
        \caption{Across-Image Insensitivity}
        \label{fig: Across_insen}
    \end{subfigure}

    \vspace{0.5em}

    % ======= subfig 2 =======
    \begin{subfigure}{\linewidth}
        \centering
        \includegraphics[width=\linewidth]{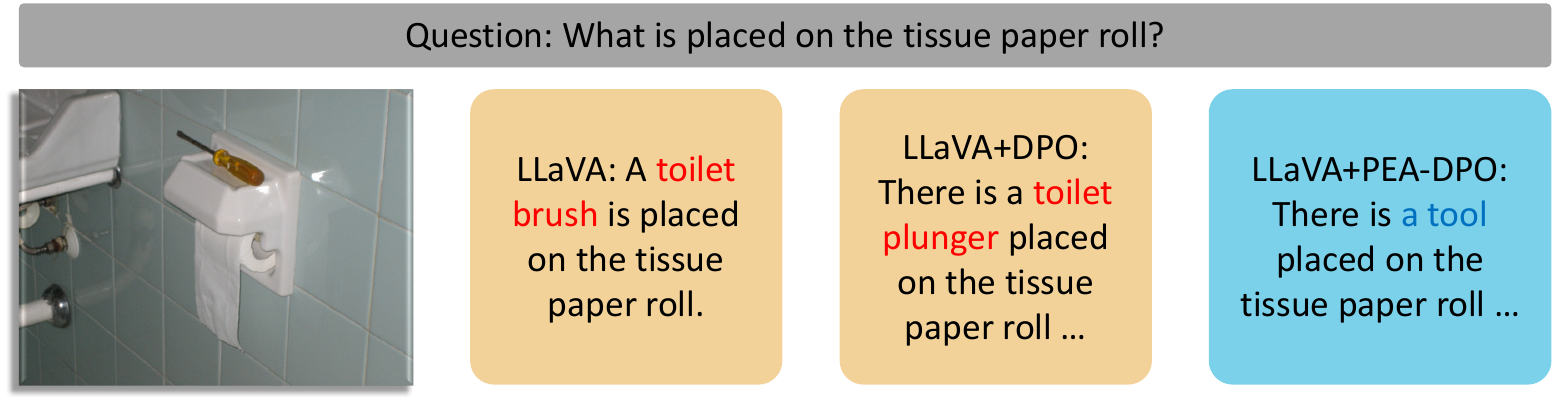}
        \vspace{-1em}
        \caption{Within-Image Insensitivity}
        \label{fig: Within_insen}
    \end{subfigure}
    \vspace{-2em}
    \caption{Illustration of two manifestations of visual insensitivity in MLLM alignment. (a) Across-image insensitivity: the model assigns nearly identical preference across distinct images (\eg outdoor reading vs. picnic), failing to capture discriminative visual context. (b) Within-image insensitivity: the model fails to discriminate semantically critical cues from irrelevant ones within the same image. (\eg confusing a screwdriver with a toilet brush or a plunger), leading to visually ungrounded responses.}
    \Description{Two examples of visual insensitivity. The first contrasts different scenes that receive nearly identical model preferences. The second highlights critical objects within one scene that the model confuses with visually unrelated objects.}
    \label{fig:visual insensitivity}
\end{figure}

However, directly applying DPO to MLLMs implicitly treats the image as \emph{conditioning} rather than a \emph{target} of preference. As a result, the model can produce responses that are weakly grounded in the visual evidence. We term this failure \textbf{visual insensitivity}. Empirically, Figure~\ref{fig:representation_analysis} shows that the representation distributions of original images (\ie chosen images) and their context-reduced counterparts (\ie rejected images, with critical visual evidence removed) largely overlap for both a base LLaVA model and its DPO-tuned version, indicating poor separation between informative and uninformative visual inputs.

To provide a theoretical foundation for these observations, we formalize visual insensitivity by decomposing the model's preference into \emph{textual} and \emph{visual} components. The analysis exposes two failure modes: (i) \textbf{Across-Image Insensitivity}, where the model assigns nearly identical preference across distinct images; and (ii) \textbf{Within-Image Insensitivity}, where the model fails to discriminate semantically critical cues from irrelevant ones within the same image. Figure~\ref{fig:visual insensitivity} illustrates both phenomena: near-identical responses across different scenes (\eg outdoor reading vs.\ picnic) and misrecognition of key objects (\eg toilet brush vs.\ generic tool).

To address these limitations, we propose PEA-DPO, which fundamentally shifts from treating images as static conditioning context to jointly optimizing response quality preferences and visual context preferences. At its core is a dual preference learning framework that optimizes two complementary signals simultaneously:
(1) response quality preference, where given the same text-image input, the model learns to distinguish high-quality responses from worse ones (standard multimodal DPO), and
(2) visual context preferences, where given the same text but different visual contexts (original vs. context-reduced images), the model should favor responses that properly utilize complete visual evidence.
We implement this dual optimization through two key components:
(1) Construction of Perception-enhanced Preference Data, wherein we generate candidate images by applying random masks to original images, then leverage CLIP embeddings to identify masked variants that eliminate the most critical visual context, creating meaningful visual grounding preference pairs; and
(2) Joint Optimization Objective, which combines both preference learning terms to simultaneously enhance response quality and visual sensitivity,
promoting differentiation between different visual contexts (cross-image sensitivity) while directing attention toward semantically critical visual elements within images (within-image sensitivity).

To evaluate the effectiveness of PEA-DPO, we conduct experiments using two sizes of LLaVA-v1.5 models ~\citep{liu2024improved}, with 7B and 13B parameters. Evaluations on MMHal Bench ~\citep{sun2023aligning}, Object HalBench ~\citep{rohrbach2018object}, and AMBER ~\citep{wang2023amber} demonstrate that PEA-DPO significantly outperforms strong commercial multimodal models such as GPT-4V ~\citep{achiam2023gpt} in multimodal scenarios. Furthermore, for both 7B and 13B models, PEA-DPO achieves strong performance on all benchmarks, highlighting its effectiveness and scalability.

\section{Related Work}
\noindent In this section, we review prior work on multimodal LLMs preference optimization from two perspectives: loss function design and preference data construction.

\noindent\textbf{Loss Function Design.}
Direct Preference Optimization (DPO)~\citep{rafailov2023direct} was originally proposed for text-only LLMs and has since been extended to multimodal settings. SymPO~\citep{liu2025mitigating} enforces theoretical consistency across modalities to ensure robustness under perturbations. AdPO~\citep{liu2025adpo} applies adversarial preference signals to strengthen multimodal models against input perturbations~\citep{zhu2026principled, zhu2026enhancing}. DAMA~\citep{lu2025damo} jointly considers data and model characteristics to adjust the optimization objective adaptively. These methods primarily optimize over response-level preferences without explicitly modeling visual preference signals. mDPO~\citep{wang2024mdpo} and V-DPO~\citep{xie2024v} take a step further by incorporating visual signals into the optimization objective.

\noindent\textbf{Preference Data Construction.}
Another line of research focuses on how to construct high-quality preference data for multimodal alignment. From the \textit{text response} perspective, LLaVA-RLHF~\citep{sun2023aligning} collects human-annotated preference labels on model responses, while RLAIF-V~\citep{yu2024rlaif} replaces human annotators with open-source MLLMs to generate AI feedback at scale. RLHF-V~\citep{yu2024rlhf} further introduces fine-grained correctional feedback to improve factual alignment. From the \textit{visual input} perspective, several recent works construct rejected images to form image-level preference pairs. MFPO~\citep{mfpo:conf/ijcai/jiang0chjh00l25} and LPOI~\citep{lpoi:conf/acl/zadehok25} generate rejected images through predefined transformations. mDPO~\citep{wang2024mdpo} uses a completely uninformative image as the rejected counterpart. OPA-DPO~\citep{yang2025mitigating} emphasizes the importance of on-policy data collection for preference pairs. CHiP~\citep{fu2025chip} proposes a hierarchical cross-modal framework that constructs preference data capturing multi-level dependencies. However, these approaches either apply coarse-grained transformations that fail to precisely remove key visual context, or rely on heavyweight models for data construction.
\section{Background}

\subsection{Background: Direct Preference Optimization in Multimodal Scenario}

To further improve the performance of MLLMs, RLHF/RLAIF requires a reward model $r(x,y,m)$ that evaluates human preference over a response $y$ given a prompt $x$ and image $m$. The standard learning objective is:
\begin{equation}\label{equ: obj}
    \max_{\pi_{\theta}}\mathbb{E}_{\mathcal{D}_y}[r(x,y,m)]-\beta\mathbb{D}_{\text{KL}}[\pi_{\theta}(\cdot\mid x,m)\mid\mid\pi_{\text{ref}}(\cdot\mid x,m)],
\end{equation}
where $\mathcal{D}$ denotes the dataset, with prompts $x$ and images $m$ sampled from the reference policy $\pi_{\text{ref}}$. The term $\mathbb{D}_{\text{KL}}$ is the KL divergence, and $\beta$ controls the strength of regularization. DPO derives a closed-form solution to Eq.~\ref{equ: obj}, revealing that the reward function can be expressed as:
\begin{equation}\label{equ: close_solu}
    r(x,y,m)=\beta\log\frac{\pi_{\theta}(y\mid x,m)}{\pi_{\text{ref}}(y\mid x,m)}+\beta\log Z(x,m),
\end{equation}
where $Z(x,m)$ is a partition function depending only on the prompt $x$ and image $m$. Incorporating this into the Bradley–Terry model~\citep{bradley1952rank}, and given a dataset of preference pairs $(y_w\succ y_l)$ under the same $(x,m)$, the optimization objective becomes:
%
% \begin{equation}\label{dpo_y_loss}
%     \begin{aligned}
%         \mathcal{L}_{\text{DPO}}&=-\mathbb{E}_{\mathcal{D}}[\log\sigma(r(x,y_w,m)-r(x,y_l,m))] \\
%         &= -\mathbb{E}_{\mathcal{D}}\bigg[\log\sigma\bigg(\beta\log\frac{\pi_{\theta}(y_w\mid x,m)}{\pi_{\text{ref}}(y_w\mid x,m)} \\
%         &\phantom{{}=-\mathbb{E}_{\mathcal{D}}\bigg[\log\sigma\bigg(}-\beta\log\frac{\pi_{\theta}(y_l\mid x,m)}{\pi_{\text{ref}}(y_l\mid x,m)}\bigg)\bigg],
%     \end{aligned}
% \end{equation}
\begin{equation}\label{dpo_y_loss}
    \begin{aligned}
        \mathcal{L}_{\text{DPO}}&=-\mathbb{E}_{\mathcal{D}}[\log\sigma(r(x,y_w,m)-r(x,y_l,m))] \\
        &= -\mathbb{E}_{\mathcal{D}}\bigg[\log\sigma\bigg(\beta\log\frac{\pi_{\theta}(y_w\mid x,m)}{\pi_{\text{ref}}(y_w\mid x,m)} -\beta\log\frac{\pi_{\theta}(y_l\mid x,m)}{\pi_{\text{ref}}(y_l\mid x,m)}\bigg)\bigg],
    \end{aligned}
\end{equation}
where $\sigma(\cdot)$ denotes the sigmoid function.

% \subsection{Background: ReLU-based Direct Preference Optimization (RePO)}
% RePO\citep{wu2025repo} is an effective and controllable variant of DPO that streamlines DPO through three key modifications: (1) it replaces the log-probability ratio in \ref{mDPO_loss} with a length-normalized log probability, thereby eliminating the reference model; (2) it substitutes the sigmoid function in \ref{mDPO_loss} with \text{ReLU} function, which filters out trivial data points during optimization and mitigates overfitting; (3) it introduces a response-based target margin $\gamma_y$ while removing the hyperparameter $\beta$ enabling controllable optimization. In the multimodal setting, the RePO loss is defined as:

% \begin{equation}\label{RePO_loss}
%     \mathcal{L}_{\text{RePO}_y}=\mathbb{E}_{\mathcal{D}_{y}}\Bigg\{\mathrm{ReLU}\Bigg[ -\bigg(
%     \frac{\log \pi_{\theta}(y_{w}|x, m_{w})}{|y_{w}|}
%     - \frac{\log \pi_{\theta}(y_{l}|x, m_{w})}{|y_{l}|}
%     - \gamma_{y} \Bigg)\Bigg]\Bigg\}.
% \end{equation}
% where $|y|$ denotes the token length of response $y$, $\gamma_y$ is a predefined margin that enforces a minimum reward gap between $y_w$ and $y_l$.

\subsection{Problem: Across-Image Insensitivity v.s. Within-Image Insensitivity}\label{problem}

In this section, we provide a theoretical analysis of visual insensitivity in multimodal DPO, which manifests in two forms: (1) across-image insensitivity and (2) within-image insensitivity. 
% \begin{definition}
%     (image-based likelihood ration) Inspired by \citep{gutmann2010noise, oord2018representation}, we define the image-based likelihood ratio in log form:
%     \begin{equation}
%         \ell_\theta(x,m;y) \triangleq \log \frac{\pi_\theta(y\mid x,m)}{\pi_\theta(y\mid x)},
%     \end{equation}
%     which quantifies the relative gain in the plausibility of response $y$ when conditioning on the image $m$ in addition to the prompt $x$.
% \end{definition}
\begin{restatable}[image-based likelihood ratio]{definition}{imagelikelihood}
Inspired by~\citep{gutmann2010noise, oord2018representation}, we define the image-based likelihood ratio in log form:
\begin{equation}\label{eq:image_likelihood}
    \ell_\theta(x,m;y) \triangleq \log \frac{\pi_\theta(y\mid x,m)}{\pi_\theta(y\mid x)},
\end{equation}
which quantifies the relative gain in the plausibility of response $y$ when conditioning on the image $m$ in addition to the prompt $x$.
\end{restatable}

% \begin{definition}
%     (response-based margin) Given a preference data $(x,m,y_w,y_l)$, the response-based margin is defined as follows:
%     \begin{equation}\label{reponse margin}
%         M^y_{\theta}(x,m)\triangleq \log\pi_\theta(y_w\!\mid x,m)-\log\pi_\theta(y_l\!\mid x,m).
%     \end{equation}
% \end{definition}
\begin{restatable}[response-level margin]{definition}{responsemargin}
Given a preference data $(x,m,y_w,y_l)$, the response-level margin is defined as follows:
\begin{equation}\label{reponse_margin}
    % M^y_{\theta}(x,m)\triangleq \log\pi_\theta(y_w\!\mid x,m)-\log\pi_\theta(y_l\!\mid x,m).
    G_{r}\triangleq \log\pi_\theta(y_w\!\mid x,m)-\log\pi_\theta(y_l\!\mid x,m).
\end{equation}
\end{restatable}
$G_{r}$ quantifies the relative preference of the policy $\pi_{\theta}$ between the preferred response $y_w$ and the dispreferred response $y_l$. $G_{r}$ can be decomposed as:
\begin{equation}\label{eq:decompose_response_margin}
    \begin{aligned}
        % M^y_{\theta}(x,m) =& \underbrace{\big[\log\pi_\theta(y_w\mid x)-\log\pi_\theta(y_l\mid x)\big]}_{\Delta_{\mathrm{text}}^\theta} \\
        % &+\underbrace{\big[\ell_\theta(x,m;y_w)-\ell_\theta(x,m;y_l)\big]}_{\Delta_{\mathrm{vis}}^\theta},
        G_{r} =& \underbrace{\big[\log\pi_\theta(y_w\mid x)-\log\pi_\theta(y_l\mid x)\big]}_{\Delta_{\mathrm{t}}} \\
        &+\underbrace{\big[\ell_\theta(x,m;y_w)-\ell_\theta(x,m;y_l)\big]}_{\Delta_{\mathrm{m}}},
    \end{aligned}
\end{equation}
where $\Delta_{\mathrm{m}}$ measures the extent to which image $m$ reinforces the preference for the preferred response. Given a pair of images $m_w,m_l$, where $m_w$ enables the prompt $x$ to better align with the chosen response $y_w$ than $m_l$. 

\begin{restatable}[Across-Image Insensitivity]{theorem}{acrossimage}
\label{theorem: across}
Suppose there exist samples $(x, m_w, m_l, y_w, y_l)$ such that
    \begin{equation}\label{hypo-1}
        G_r(m_w)-G_r(m_l)\leq\delta\quad(\delta\to0),
    \end{equation}
    it implies the relation:
    \begin{equation}
        \Delta_{\mathrm{m}}(m_w)-\Delta_{\mathrm{m}}(m_l)\leq\delta\quad(\delta\to0).
    \end{equation}
\end{restatable}
\noindent\textit{Intuitive explanation.} The model cannot effectively distinguish the impact of $m_w$ versus $m_l$ on the responses, a phenomenon we term \textbf{Across-Image Insensitivity}.

\begin{figure*}[htbp]
    \centering
    \includegraphics[width=0.95\linewidth]{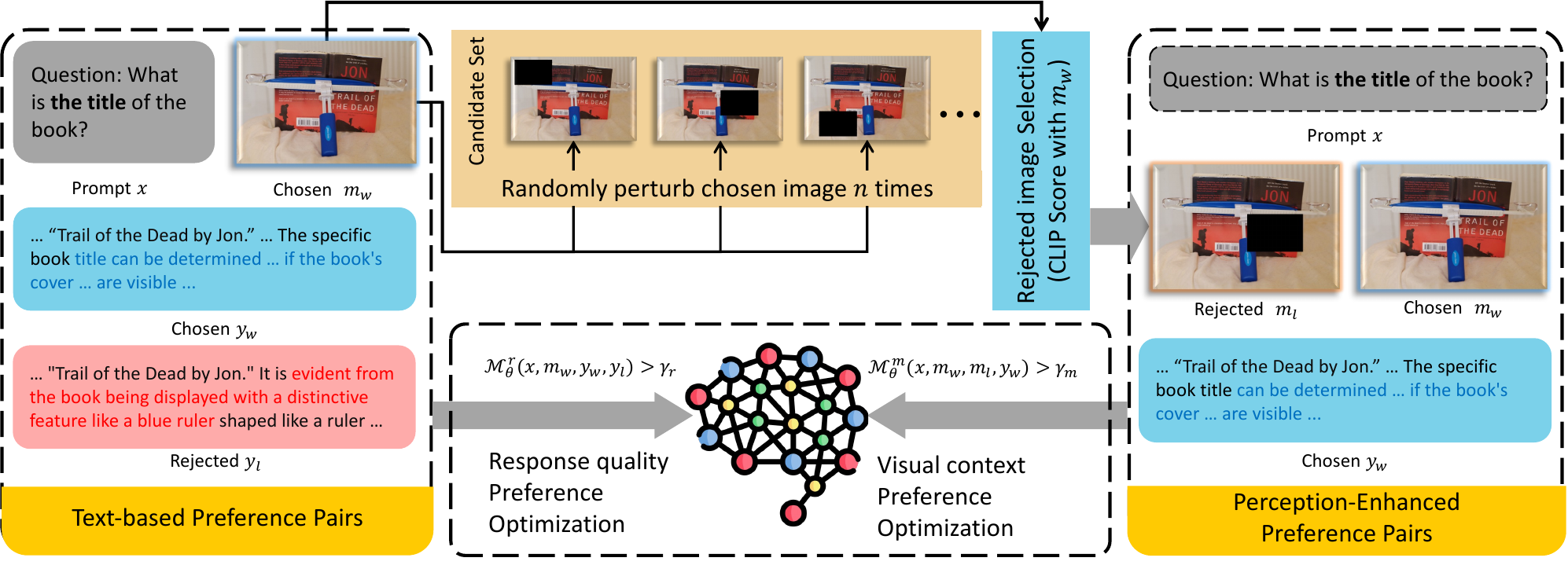}
    \vspace{-12pt}
    \caption{Overview of \textbf{PEA-DPO}. \textit{Left}: Original text-based preference pairs.
    \textit{Top}: Construction process of perception-enhanced preference pairs, where critical key visual context is removed from images using a CLIP-based approach. \textit{Right}: The resulting perception-enhanced preference pairs. \textit{Bottom}: Joint optimization text-based and perception-enhanced preferences.
    \Description{The PEA-DPO pipeline starts with an image, prompt, preferred response, and rejected response. It masks candidate image regions, uses CLIP similarity to select a context-reduced image, constructs visual-context preference pairs, and jointly optimizes textual response and visual-context preferences.}
    }
    \label{fig:overview}
\end{figure*}

\begin{restatable}[Within-Image Insensitivity]{theorem}{withinimage}
\label{theorem: within}
Suppose there exist samples $(x,m_l, y)$ such that
\begin{equation}\label{hypo-2}
    \big|\ell_\theta(x,m_l;y)\big|\le \varepsilon \quad (\varepsilon\to0),
\end{equation}
and that the model exhibits \textbf{Across-Image Insensitivity}. Then the following holds for $m_w$:
\begin{equation}
    \big|\ell_\theta(x,m_w;y_w)-\ell_\theta(x,m_w;y_l)\big|\le \delta + 2\varepsilon \quad(\delta,\varepsilon\to0).
\end{equation}
\end{restatable}
\noindent\textit{Intuitive explanation.} This observation suggests that the model is unable to reliably generate the correct response even when conditioned on $m_w$. We refer to this phenomenon as \textbf{Within-Image Insensitivity}. Complete proofs are provided in Appendix~\ref{proof}.

\section{Method}

In summary, both types of issues arise from visual insensitivity. To address these limitations, we introduce PEA-DPO, as illustrated in Figure~\ref{fig:overview}. It consists of two key components:
(1) Construction of Perception-Enhanced Preference Data, where rejected images $m_l$ are generated by removing critical visual context using a CLIP-based approach; and
(2) Joint Optimization Objective, which learns response quality preferences and visual sensitivity, simultaneously optimizing over response-level and image-level preferences, thereby aligning with human value preferences while enhancing visual sensitivity.

\subsection{Construction of Perception-Enhanced Preference Data}
We begin with the chosen image $m_w$ (\ie the original image) and apply a random mask of fixed proportion to generate a perturbed image $m_p$:
\begin{equation}
m_p=m_w\odot(1-\mathbf{P}),    
\end{equation}
where $\mathbf{P}$ denotes a random binary mask and $\odot$ represents the Hadamard product. Repeating this process $n$ times yields a candidate set of perturbed images:
\begin{equation}
    \mathcal{M}_p=\{m_{p}^k\}_{k=1}^{n},
\end{equation}
where $m^k_p$ denoting the $k$-th perturbed image. Since the perturbations are random, each perturbed image retains different portions of the original visual information. To quantify how much visual context is lost relative to the chosen image $m_w$, we compute semantic similarity using CLIP ~\citep{radford2021learning} embeddings. Let $\mathbf{v}_{w}=f_{\mathrm{CLIP}}(m_{w})$ and $\mathbf{v}^k_{p}=f_{\mathrm{CLIP}}(m^k_{p})$ denote the $\ell_2$-normalized embeddings from the CLIP image encoder $f_{\mathrm{CLIP}}(\cdot)$. The similarity score is defined as:
\begin{equation}
    s_k = \cos(\mathbf{v}_w, \mathbf{v}^k_p) = \frac{\mathbf{v}_w^\top \mathbf{v}^k_p}{\|\mathbf{v}_w\|_2 \, \|\mathbf{v}^k_p\|_2},
\end{equation}
where $s_k\in [-1,1]$ measures semantic similarity. Intuitively, given that all perturbations have equal mask size, a lower similarity score indicates that critical visual context has been removed. Finally, we select the perturbed image with the lowest similarity as the rejected image: 
\begin{equation}
    m_l=\arg\min_{m^k_p\in\mathcal{M}_p}s_{k},
\end{equation}
where $m_l$ corresponds to the image with its key visual context removed. Optimizing over the perception-enhanced preference data $(x,m_l,m_w,y_w)$ encourages the model to leverage critical visual cues, thereby improving its ability to generate preferred responses grounded in visual evidence.

\subsection{Joint Optimization of Response quality and Visual context Preferences}

After constructing the Perception-Enhanced Preference Data, we obtain a dataset $\mathcal{D}_{m}=(x,m_w,m_l,y_w)$. Building upon the standard multimodal DPO, we replace the response quality preference data $\mathcal{D}=(x,m_w,y_w,y_l)$ with $\mathcal{D}_{m}$ in Eq.~\ref{dpo_y_loss}. Let $h(x,m,y)=\beta\log\frac{\pi_{\theta}(y\mid x,m)}{\pi_{\text{ref}}(y\mid x,m)}$. This yields Visual context Preference Optimization (VPO) Objective, formulated as:
\begin{equation}\label{dpo_m_loss}
\begin{aligned}
    \mathcal{L}_{\text{VPO}} &= -\mathbb{E}_{\mathcal{D}_m}\bigg[\log\sigma\big(h(x,m_{w},y_{w}) - h(x,m_{l},y_{w})\big)\bigg].
\end{aligned}
\end{equation}

Combining this with the standard multimodal DPO (\ie Response quality Preference Optimization, RPO) in Eq.~\ref{dpo_y_loss}, we obtain the PEA-DPO objective:
\begin{equation}\label{loss}
    \begin{aligned}
        \mathcal{L}_{\text{PEA-DPO}}&=\mathcal{L}_{\text{RPO}}+\alpha\cdot\mathcal{L}_{\text{VPO}}\\
        &=-\mathbb{E}_{\mathcal{D}}\bigg[\log\sigma\bigg(h(x,m_{w},y_{w})-h(x,m_{w},y_{l})\bigg)\bigg] \\
        &\phantom{{}=}-\mathbb{E}_{\mathcal{D}_m}\bigg[\log\sigma\bigg(h(x,m_{w},y_{w})-h(x,m_{l},y_{w})\bigg)\bigg],
    \end{aligned}
\end{equation}
where $\alpha$ is a weighting hyperparameter. This joint objective enables MLLMs to align with human preferences by simultaneously leveraging textual and critical visual modalities. 

To further reduce computational overhead and enable controllable preference optimization, inspired by the design of RePO ~\citep{wu2025repo}, we introduce three modifications to $\mathcal{L}_{\text{PEA-DPO}}$: (1) Replace log-probability ratios in Eq.~\ref{loss} with length-normalized log-probabilities, thereby eliminating the need for a reference model; (2) Replace the sigmoid function with a ReLU activation, which filters out trivial data points and prevents overfitting; (3) Introduce target margin $\{\gamma_r,\gamma_m\}$ and remove the temperature parameter $\beta$, enabling a controllable optimization process. Let $h_{m}(x,m,y)=\frac{\log \pi_{\theta}(y|x, m)}{|y|}$. The modified PEA-DPO loss function is then given as:
\begin{equation}\label{mloss}
    \small
    \begin{aligned}
        &\mathcal{L}_{\text{mPEA-DPO}}=\mathcal{L}_{\text{mRPO}}+\alpha\cdot\mathcal{L}_{\text{mVPO}} \\
        &=\mathbb{E}_{\mathcal{D}}\bigg\{\mathrm{ReLU}\bigg[ -\big( h_{m}(x,m_{w},y_{w}) 
        - h_{m}(x,m_{w},y_{l}) - \gamma_{r} \big)\bigg]\bigg\}\\
        &+\alpha\mathbb{E}_{\mathcal{D}_{m}}\bigg\{\mathrm{ReLU}\bigg[ -\big( h_{m}(x,m_{w},y_{w}) - h_{m}(x,m_{l},y_{w}) - \gamma_{m}\big)\bigg]\bigg\},
    \end{aligned}
\end{equation}
where $|y|$ denotes the number of tokens in response $y$, and $\{\gamma_r,\gamma_m\}$ are the target reward margins, enforcing a minimum separation between preferred and rejected responses in both response-based preferences $\{y_w,y_l\}$ and image-based preferences $\{m_w,m_l\}$.
% We provide detailed analyses in the Appendix~\ref{theorety_and_experiment_of_mpeadpo} explaining why mPEA-DPO can both eliminate across‑ and within‑image insensitivity

% We provide both theoretical and empirical analyses in the Appendix~\ref{theorety_and_experiment_of_mpeadpo} explaining why mPEA-DPO (jointly optimizing response quality and visual context preferences) can eliminate visual insensitivity, \ie across‑image and within‑image insensitivity.
\section{Theoretical Analysis: How mPEA-DPO Mitigates Visual Insensitivity}

We now establish the theoretical connection between the mPEA-DPO objective and the two forms of visual insensitivity identified in Section~\ref{problem}. Starting from definitions in Eq.~\ref{eq:image_likelihood} and Eq.~\ref{reponse_margin}, we show that each failure mode corresponds to a specific margin being small, and that mPEA-DPO is designed to directly enlarge both margins.

\subsection{Image-Level Margin and Across-Image Insensitivity.}
We define the image-level margin as:
\begin{equation}
    G_m = \log \pi_\theta(y_w \mid x, m_w) - \log \pi_\theta(y_w \mid x, m_l),
\end{equation}
which measures how much the model's confidence in the preferred response $y_w$ increases when replacing the context-reduced image $m_l$ with the correct image $m_w$. According to the definition of $\Delta_\text{m}$ in Eq.~\ref{eq:decompose_response_margin} and the image-based likelihood ratio in Eq.~\ref{eq:image_likelihood}, we can derive (see Appendix~\ref{theorety_and_experiment_of_mpeadpo} for full details):
\begin{align}
    &\Delta_\text{m}(x, m_w) - \Delta_\text{m}(x, m_l) \nonumber \\
    =& \underbrace{\log \pi_\theta(y_w \mid x, m_w) - \log \pi_\theta(y_w \mid x, m_l)}_{\text{image-level margin } G_m} \nonumber \\
    &+ \underbrace{\log \pi_\theta(y_l \mid x, m_l) - \log \pi_\theta(y_l \mid x, m_w)}_{\text{secondary term}}.
\end{align}
When the model exhibits Across-Image Insensitivity (Theorem~\ref{theorem: across}), both sides are bounded by $\delta \to 0$, implying that $G_m$ is near zero. That is, the model's confidence in the preferred response barely changes regardless of whether the correct or context-reduced image is provided. The secondary term captures how the dispreferred response shifts across images, but improving $G_m$ has a more direct impact on performance as it explicitly boosts the probability of the correct output under the chosen image.

\subsection{Response-Level Margin and Within-Image Insensitivity.}
% We define the response-level margin as:
% \begin{equation}
%     G_r = \log \pi_\theta(y_w \mid x, m_w) - \log \pi_\theta(y_l \mid x, m_w),
% \end{equation}
% which measures the model's ability to discriminate the preferred response $y_w$ from the dispreferred $y_l$ when conditioned on the correct image $m_w$.
Similarly, expanding using the image-based likelihood ratio yields:
\begin{align}
    &\ell_\theta(x, m_w; y_w) - \ell_\theta(x, m_w; y_l) \nonumber \\
    =& \underbrace{\log \pi_\theta(y_w \mid x, m_w) - \log \pi_\theta(y_l \mid x, m_w)}_{\text{response-level margin } G_r} \nonumber \\
    &+ \underbrace{\log \pi_\theta(y_l \mid x) - \log \pi_\theta(y_w \mid x)}_{\text{text-only bias}}.
\end{align}
When the model exhibits Within-Image Insensitivity (Theorem~\ref{theorem: within}), this quantity is bounded by $\eta = \delta + 2\varepsilon \to 0$, implying that $G_r$ is near zero. The model fails to distinguish the preferred from the dispreferred response even with the correct visual evidence. The text-only bias term reflects inherent language priors without visual input; improving $G_r$ is more critical as it directly enhances visual grounding.

\subsection{mPEA-DPO Directly Enlarges Both Margins.}
With the two failure modes characterized by small $G_m$ and $G_r$ respectively, we show that the mPEA-DPO objective is designed to directly enlarge both. Revisiting Eq.~\ref{mloss}:
\begin{equation}
    \begin{split}
    % &\mathcal{L}_\text{mPEA-DPO} \\
    % =& \underbrace{\mathbb{E}_{\mathcal{D}} \left\{ \text{ReLU} \left[ -\left( \frac{\log \pi_\theta(y_w|x,m_w)}{|y_w|} - \frac{\log \pi_\theta(y_l|x,m_w)}{|y_l|} - \gamma_r \right) \right] \right\}}_{\mathcal{L}_\text{mRPO}: \text{ increases } G_r \text{, mitigates Within-Image Insensitivity}} \nonumber \\
    % &+ \alpha \cdot \underbrace{\mathbb{E}_{\mathcal{D}_m} \left\{ \text{ReLU} \left[ -\left( \frac{\log \pi_\theta(y_w|x,m_w)}{|y_w|} - \frac{\log \pi_\theta(y_w|x,m_l)}{|y_w|} - \gamma_m \right) \right] \right\}}_{\mathcal{L}_\text{mVPO}: \text{ increases } G_m \text{, mitigates Across-Image Insensitivity}}.
    &\mathcal{L}_\text{mPEA-DPO} \\
    =& \underbrace{\mathbb{E}_{\mathcal{D}} \left\{ \text{ReLU} \left[ -\left(h(x,m_w,y_w) - h(x,m_w,y_l) - \gamma_r \right) \right] \right\}}_{\mathcal{L}_\text{mRPO}: \text{ increases } G_r \text{, mitigates Within-Image Insensitivity}} \nonumber \\
    &+ \alpha \cdot \underbrace{\mathbb{E}_{\mathcal{D}_m} \left\{ \text{ReLU} \left[ -\left( h(x,m_w,y_w) - h(x,m_l,y_w) - \gamma_m \right) \right] \right\}}_{\mathcal{L}_\text{mVPO}: \text{ increases } G_m \text{, mitigates Across-Image Insensitivity}}.
    \end{split}
\end{equation}
The $\mathcal{L}_\text{mVPO}$ term enforces $h(x,m_w,y_w) - h(x,m_l,y_w) > \gamma_m$, directly increasing the image-level margin $G_m$ and enhancing the model's discrimination between the correct and context-reduced images. The $\mathcal{L}_\text{mRPO}$ term enforces $h(x,m_w,y_w) - h(x,m_w,y_l) > \gamma_r$, directly increasing the response-level margin $G_r$ and enhancing the model's ability to favor the preferred response under the correct image. Since $\mathcal{L}_\text{mPEA-DPO}$ is a linear combination of both terms, it provably addresses Across-Image and Within-Image Insensitivity simultaneously. Following SimPO~\citep{meng2024simpo}, we normalize log-probabilities by response length to prevent length bias. Full derivations and an empirical validation showing how $G_m$ and $G_r$ shift before and after training are provided in Appendix~\ref{theorety_and_experiment_of_mpeadpo}.

\section{Experiments}

\begin{table*}[t]
\caption{\textbf{Main results} of LLaVA-v1.5-7B and LLaVA-v1.5-13B trained with different preference optimization objectives. 
We report overall score (Score) and hallucination rate (Hal.) on MMHalBench, CHAIR scores at both response and object levels on Object HalBench, along with CHAIR scores (C.), object coverage (cover.), hallucination rate (Hal.), and cognition (Cog.) on AMBER. The best result for each metric in each group is in bold. We have carefully followed the publicly available code and released checkpoints to reproduce these results, aiming to provide a fair comparison. $^\dagger$ indicates results obtained using the official API, $^\ddagger$ indicates results reproduced using the authors’ released code, and $^\sharp$ indicates results produced from the authors’ provided checkpoints.}
% For completeness, we further report additional results obtained with a range of multimodal LLMs, preference datasets, and training objectives, even though these results are not directly comparable.
\vspace{-1em}
\centering
\setlength{\tabcolsep}{2.5pt}
\begin{tabular}{lcccccccc}\toprule
 &\multicolumn{2}{c}{MMHalBench} &\multicolumn{2}{c}{Object HalBench} &\multicolumn{4}{c}{AMBER} \\ \cmidrule(lr){2-3} \cmidrule(lr){4-5} \cmidrule(lr){6-9}
&Score $\uparrow$ &Hal. $\downarrow$ &CHAIR$_{s}$ $\downarrow$&CHAIR$_{i}$ $\downarrow$&C. $\downarrow$&Cover. $\uparrow$ &Hal. $\downarrow$&Cog. $\downarrow$\\\midrule
GPT-4V~\citep{achiam2023gpt}$^\dagger$ &3.49 & 0.28 &13.6 &7.3 &4.6 &67.1 &30.7 &2.6 \\
Gemini-2.5-pro~\citep{comanici2025gemini}$^\dagger$ & 3.55 & 0.28 & 12.0 & 8.2 & 9.5 & 78.0 & 75.1 & 5.2 \\ \midrule
\rowcolor{lightgray} \multicolumn{9}{c}{\textbf{\textit{7B MLLMs}}} \\ \midrule
LLaVA-v1.5-7B~\citep{liu2024improved} & 2.11 & 0.54 & 53.6 & 25.2 & 7.8 & 51.0 & 36.4 & 4.2 \\
+ HACL~\citep{jiang2024hallucination}$^\ddagger$ &2.13 &0.50 &- &- &- &- &- &- \\
+ OPERA~\citep{huang2024opera}$^\ddagger$& 2.15 & 0.54 & 45.1 & 22.3  &- &- &- &- \\
+ VCD~\citep{leng2024mitigating}$^\ddagger$& 2.12 & 0.54 & 48.8 & 24.3 &- &- &- &- \\
+ EOS~\citep{yue2024less}$^\ddagger$ &2.03 & 0.59 & 40.3 & 17.8 & 5.1 & 49.1 & 22.7 & 2.0 \\
+ POVID~\citep{zhou2024aligning}$^\ddagger$ &2.08 &0.56 &48.1 &24.4 &- &- &- &- \\
+ LLaVA-RLHF~\citep{sun2023aligning}$^\sharp$ &1.88 &0.71 &58.0 &15.6 &9.7 & \textbf{53.2} &46.6 &5.3 \\ 
+ HA-DPO~\citep{zhao2023beyond}$^\sharp$ &1.97 &0.60 & 39.9 &19.9 &6.7 & 49.8 &30.9 &3.3 \\
+ HALVA~\citep{sarkar2024mitigating}$^\sharp$ &2.25 &0.54 &- &- &6.6 & \underline{53.0} &32.2 &3.4 \\
+ mDPO~\citep{wang2024mdpo}$^\ddagger$ &2.39 &0.54 &35.7 &9.8 &4.4 &52.4 & 24.5 &2.4 \\
+ RLAIF-V~\citep{yu2024rlaif}$^\sharp$ &3.00 & \underline{0.38} &16.0 & \underline{3.7} &3.0 &50.4 &16.2 &1.0 \\
+ OPA-DPO~\citep{yang2025mitigating}$^\sharp$ & \underline{2.83} & 0.45 &13.0 &4.3 & \underline{2.2} &47.9 & \underline{11.6} & \underline{0.9} \\ 
+ DAMA~\citep{lu2025damo}$^\sharp$ &2.76 &0.41 & \underline{10.3} &5.9 &3.0 &48.3 &14.8 &1.2 \\
+ \textbf{mPEA-DPO}  &\textbf{3.02} &\textbf{0.36} &\textbf{4.3} &\textbf{3.2} &\textbf{1.9} &46.7 &\textbf{10.3} &\textbf{0.6} \\ \midrule

\rowcolor{lightgray} \multicolumn{9}{c}{\textbf{\textit{13B MLLMs}}} \\ \midrule
LLaVA-v1.5-13B~\citep{liu2024improved} & 2.42 & - & 46.3 & 22.6 & 7.8 & 51.0 & 36.4 & 4.2  \\
+ LLaVA-RLHF~\citep{sun2023aligning}$^\sharp$ &2.27 &0.64 &44.7 &11.8 &7.7 & \underline{52.3} &38.6 &4.0 \\ 
+ RLHF-V~\citep{yu2024rlhf}$^\sharp$ &2.81 &0.49 &12.2 &7.5 &6.3 &46.1 &25.1 &2.1 \\
+ HALVA~\citep{sarkar2024mitigating}$^\sharp$ &2.58 &0.45 &- &- &6.4 & \textbf{52.6} &30.4 &3.2 \\
+ OPA-DPO~\citep{yang2025mitigating}$^\sharp$ & \underline{3.07} & \underline{0.39} &16.33 &5.5 & \textbf{2.4} & 48.3 &\textbf{12.8} & \textbf{0.8} \\ 
+ DAMA~\citep{lu2025damo}$^\sharp$ &2.89 &0.43 & \underline{7.7} & \underline{4.9} &3.0 &50.5 &14.1 &0.9 \\
+ \textbf{mPEA-DPO}  &\textbf{3.16} & \textbf{0.31} &\textbf{4.3} &\textbf{2.7} & \underline{2.9} &50.0 &\textbf{12.8} &\textbf{0.8} \\

\bottomrule
\end{tabular}
\label{tab: main}
\end{table*}

\subsection{Experimental Setup}

\subsubsection{Models.}
We evaluate mPEA-DPO on two MLLMs with different parameter scales: LLaVA-v1.5-7B and LLaVA-v1.5-13B~\citep{liu2024improved}, both equipped with a CLIP ViT-L-336px vision encoder. The 7B model is built upon Vicuna-7B as its LLM backbone, while the 13B model utilizes Vicuna-13B as its LLM backbone. Both models are first pretrained on 558K image–text pairs datasets and then fine-tuned on 665K instruction-following instances.

\subsubsection{Training Data.}
Following~\citep{lu2025damo}, we utilize the preference data of the LLaVA-1.5 model released by~\citep{yu2024rlaif}, where the language-based preference is annotated by the open-source LLaVA-NeXT-34B model. Specifically, the dataset comprises 22K preference instances, with 13K images used for training. To construct perception-enhanced preference data, we generate rejected images by masking informative regions in these 13K images.

\subsubsection{Baselines.}

We report results across three categories of multimodal alignment approaches, while noting that direct comparison is non-trivial due to differences in base models, preference data, and alignment strategies. Specifically: 
(1) \textbf{Hallucination-specific baselines}: including VCD~\citep{leng2024mitigating}, OPERA~\citep{huang2024opera}, HALC~\citep{jiang2024hallucination}, and EOS~\citep{yue2024less},
(2) \textbf{RLHF/RLAIF-based baselines}: including POVID~\citep{zhou2024aligning}, LLaVA-RLHF~\citep{sun2023aligning}, HALVA~\citep{sarkar2024mitigating}, RLHF-V~\citep{yu2024rlhf}, HA-DPO~\citep{zhao2023beyond}, HSA-DPO~\citep{xiao2024detecting}, RLAIF-V~\citep{yu2024rlaif}, mDPO~\citep{wang2024mdpo}, OPA-DPO~\citep{yang2025mitigating}, and DAMA~\citep{lu2025damo}.
(3) \textbf{Proprietary baseline}: GPT-4V~\citep{achiam2023gpt}, which we use as a robust reference to compare the performance between open-source and proprietary commercial models. Gemini-2.5-Pro~\citep{comanici2025gemini}, which is also included in our comparison, given its status as a strong and up-to-date commercial baseline.

\subsubsection{Benchmarks.}
(1) \textbf{Object HalBench}~\citep{rohrbach2018object} is a widely adopted benchmark for evaluating object hallucination, focusing on detailed image descriptions of visual content. Following the protocol in~\citep{yu2024rlaif}, we evaluate on 300 instances and report hallucination rates at both the response-level ($\text{CHAIR}_{S}$) and object-level($\text{CHAIR}_{I}$).
(2) \textbf{AMBER}~\citep{wang2023amber} provides a multi-dimensional evaluation of hallucination in MLLMs. Using its generative task with 1K samples, we report CHAIR scores, object coverage, hallucination rates, and alignment with human cognition.
(3) \textbf{MMHal-Bench}~\citep{sun2023aligning} is a question-answering benchmark comprising 96 image–question pairs across 12 object categories and 8 question types. Following the setup of~\citep{yu2024rlaif}, we assess both overall response quality (scored from zero to six) and hallucination rate, as judged by GPT-4.

% \vspace{-10pt}

\subsubsection{Implementation Details.} 
In our experiments, we construct perception-enhanced preference data by masking $9\%$ of each image, repeated $n=30$ times. We employ LLaVA-1.5-7B and LLaVA-1.5-13B as backbone models, and perform full-parameter fine-tuning for 5,000 steps. Specifically, the batch size is set to 32 for the 7B model and 16 for the 13B model. In our objective Eq.~\ref{mloss}, we set the hyperparameters as $\gamma_{r}=1.5$, $\gamma_{m}=4.5$ and $\alpha=0.2$. All experiments are conducted using 4 NVIDIA A100 80GB GPUs.

\subsection{Main Results}
The experimental results of applying mPEA-DPO to LLaVA-v1.5-7B and LLaVA-v1.5-13B across various hallucination benchmarks are presented in Table~\ref{tab: main}. The main findings are summarized as follows: (1) mPEA-DPO significantly reduces the hallucinations of the 7B and 13B models. Compared with the 7B (13B) base model, mPEA-DPO reduces the hallucination rate in MMHalBench by 33\%, the response-level and object-level hallucination rates in Object HalBench by 92\% and 87\% respectively, and the CHAIR, hallucination rate, and human cognition in AMBER by 76\%, 72\%, and 86\% respectively. (2) Compared with existing MLLMs alignment methods, for LLaVA-v1.5-13B, mPEA-DPO achieves the best performance on 75.0\% of the hallucination metrics, and for LLaVA-v1.5-7B, it increases to 87.5\%. (3) However, these enhancements lead to a slight compromise in coverage metrics. This indicates that models trained with mPEA-DPO tend to adopt a slightly conservative strategy, avoiding uncertain assertions. Such a strategy enhances the credibility of the responses but may overlook some ambiguous details, which requires a trade-off.

\subsection{Ablation}

\subsubsection{Impact of Component Combination.} To evaluate the contribution of each component in mPEA-DPO and the effect of their combinations, we conducted a comprehensive ablation study on mPEA-DPO based on LLaVA. The experimental results are shown in Table~\ref{tab:module_ablation}. The main observations are as follows: (1) Both Response quality Preference Optimization (mRPO) and Visual context Preference Optimization (mVPO) are effective. On the  MMHalBench and AMBER datasets, both mRPO (mPEA-DPO-$\mathcal{L_{\textrm{mRPO}}}$) and mVPO (mPEA-DPO-$\mathcal{L_{\textrm{mVPO}}}$) outperform the base model.  This suggests that: (a) during optimization process, Response quality Preference Optimization (RPO) enables the model to concentrate on more challenging data points and mitigates overfitting in DPO; (b) the introduction of Visual context Preference Optimization (mVPO) enhances the model’s alignment between the image and text. (2) The combination of visual context preference optimization and response quality preference optimization strategies makes preference optimization the most powerful.

\begin{table}[htbp]
  \centering
  \caption{The ablation results of mPEA-DPO based on LLaVA.  
  Values in \textbf{bold} denote the best performance. }
  \vspace{-1em}
  % \footnotesize
  % \renewcommand\tabcolsep{2pt}
  
  \begin{tabular}{lcccc}
    \toprule
    \multirow{2}[2]{*}{Model} & \multicolumn{2}{c}{MMHalBench} & \multicolumn{2}{c}{AMBER} \\
    \cmidrule(lr){2-3}\cmidrule(lr){4-5}
          & Score$\uparrow$ & Hal.$\downarrow$ & C.↓ & Cog.↓ \\
    \midrule
    LLaVA-v1.5-7B   & 2.11 & 0.54  & 7.8  & 4.2 \\
    \midrule
    mPEA-DPO & \textbf{3.02}  & \textbf{0.36}  & \textbf{1.9}  & \textbf{0.6} \\
    -$\mathcal{L_{\textrm{mRPO}}}$  & 2.44  & 0.45  & 4.2  & 0.9 \\
    -$\mathcal{L_{\textrm{mVPO}}}$  & 2.81  & 0.38  & 2.5  & 0.9 \\
    \bottomrule
  \end{tabular}%
  \label{tab:module_ablation}
\end{table}

\subsubsection{Impact of Visual context Preference Optimization.} 
\begin{figure}[htbp]
    \centering
    \begin{subfigure}{0.36\linewidth}
        \centering
        \includegraphics[width=\linewidth]{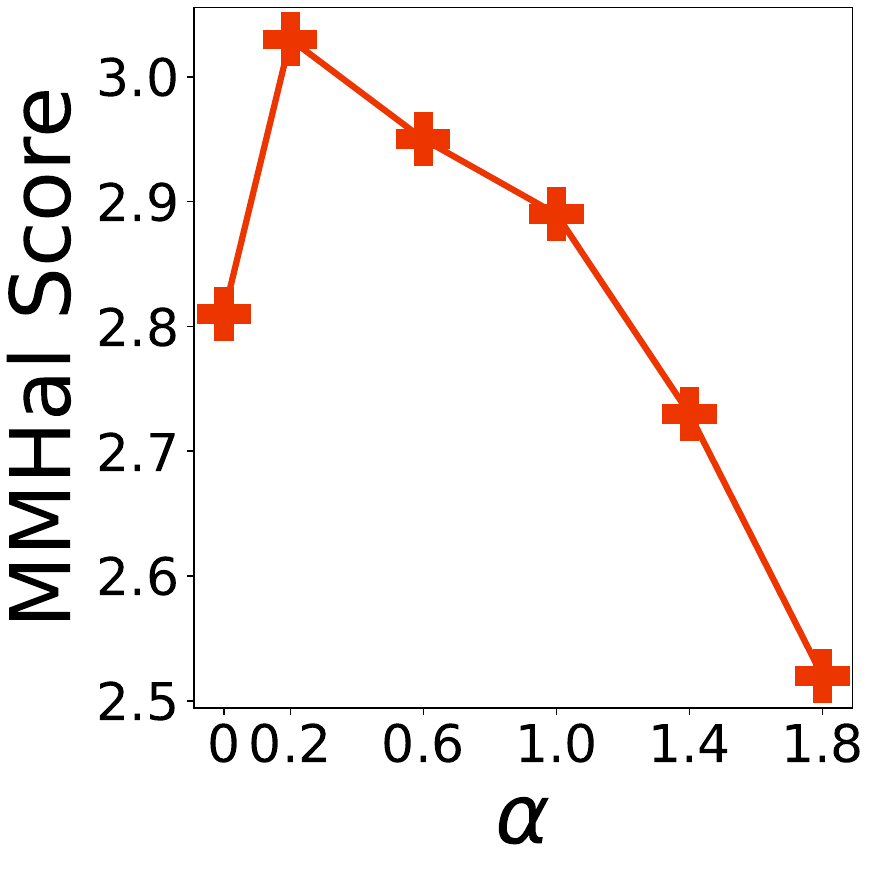}
        \vspace{-2em}
        \caption{MMHal Score}
    \end{subfigure}
    % \hfill
    \hspace{2em}
    \begin{subfigure}{0.36\linewidth}
        \centering
        \includegraphics[width=\linewidth]{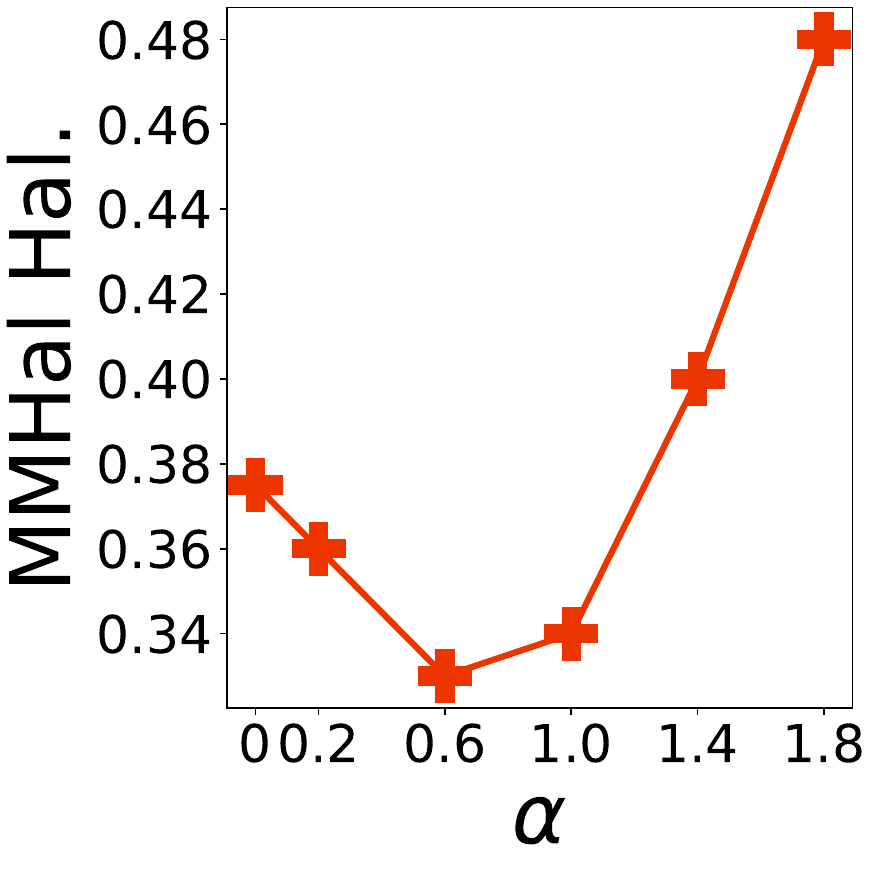}
        \vspace{-2em}
        \caption{MMHal Hal.}
    \end{subfigure}

    \vspace{0.5em}

    \begin{subfigure}{0.36\linewidth}
        \centering
        \includegraphics[width=\linewidth]{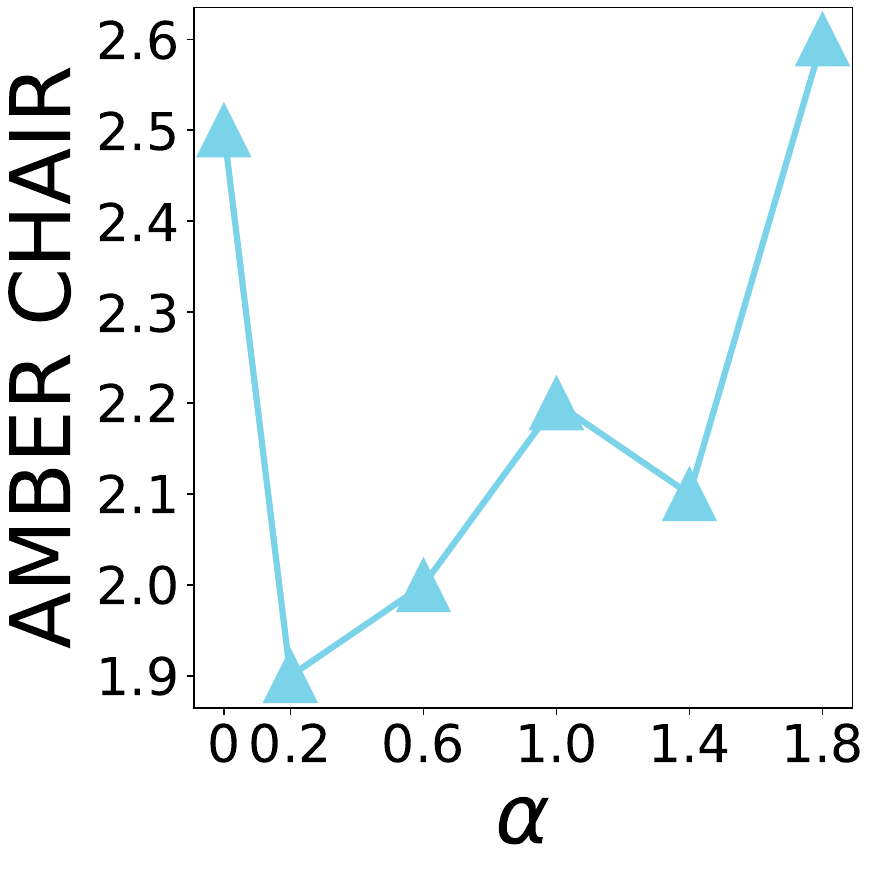}
        \vspace{-2em}
        \caption{AMBER C.}
    \end{subfigure}
    % \hfill
    \hspace{2em}
    \begin{subfigure}{0.36\linewidth}
        \centering
        \includegraphics[width=\linewidth]{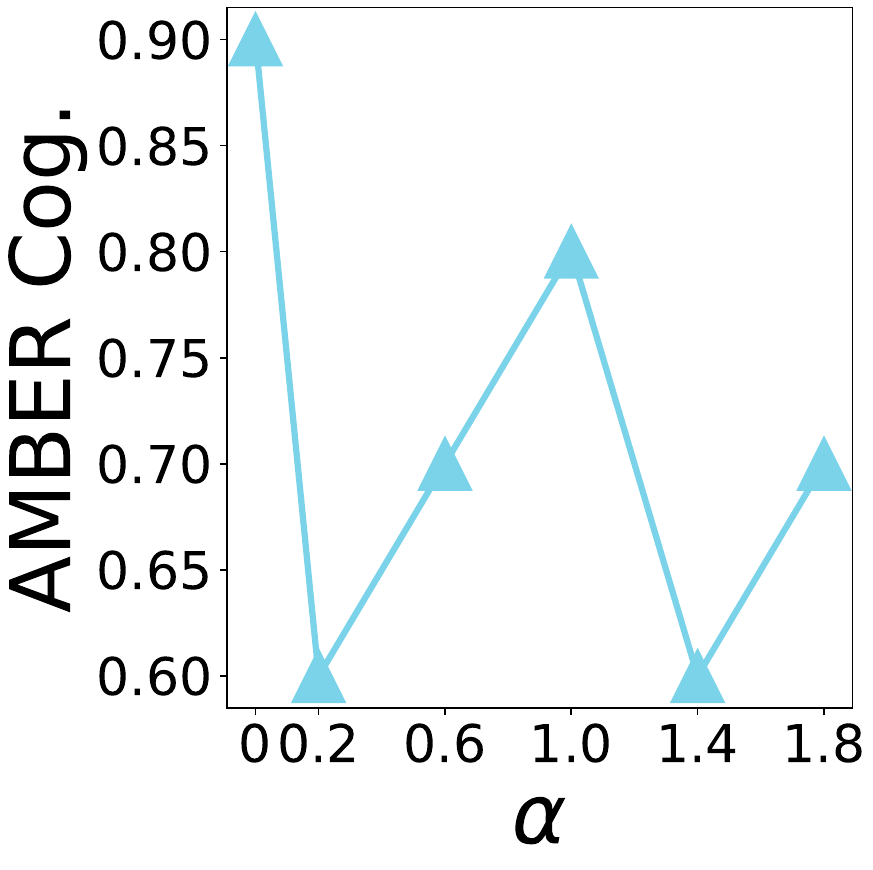}
        \vspace{-2em}
        \caption{AMBER Cog.}
    \end{subfigure}
    \vspace{-1em}
    \caption{Results of mPEA-DPO evaluated on the MMHalBench and AMBER dataset with different choices of weight $\alpha$ to control the strength of Visual context Preference Optimization. Findings: when $\alpha=0.2$, the best performance of the Score, CHAIR and Hallucination Rate metric is achieved on MMHal-Bench and AMBER based on LLaVA.}
    \Description{Four line charts show MMHal score, MMHal hallucination rate, AMBER CHAIR, and AMBER cognitive hallucination as the visual-context preference weight changes. The best overall tradeoff occurs at alpha equal to 0.2.}
    \label{fig:strength_alpha}
\end{figure}

Visual context Preference Optimization (mVPO) forces models to make preference judgements based on critical visual context. Here, we discuss the impact of its weight. We fully consider Response quality Preference Optimization (mRPO) since its global textual semantics by setting its parameter to 1 in Eq.~\ref{mloss}. As for Visual context Preference Optimization, given its crucial role in enhancing MLLM's attention to critical visual context, we fully explore the range of its weight $\alpha$ (as shown in Eq.~\ref{mloss}). For the results of Figure~\ref{fig:strength_alpha}, we observe that the best performance was achieved when $\alpha=0.2$ for LLaVA frameworks.

\subsection{Impact of rejected image construction strategy}

The quality of visual preference data depends on the rejected image quality and its gap from the chosen images. In this section, we compare the impact of our proposed strategy for constructing rejected images and various existing strategies on preference optimization.

\begin{figure}[htbp]
    \centering
    \begin{subfigure}{0.31\linewidth}
        \centering
        \includegraphics[width=\linewidth]{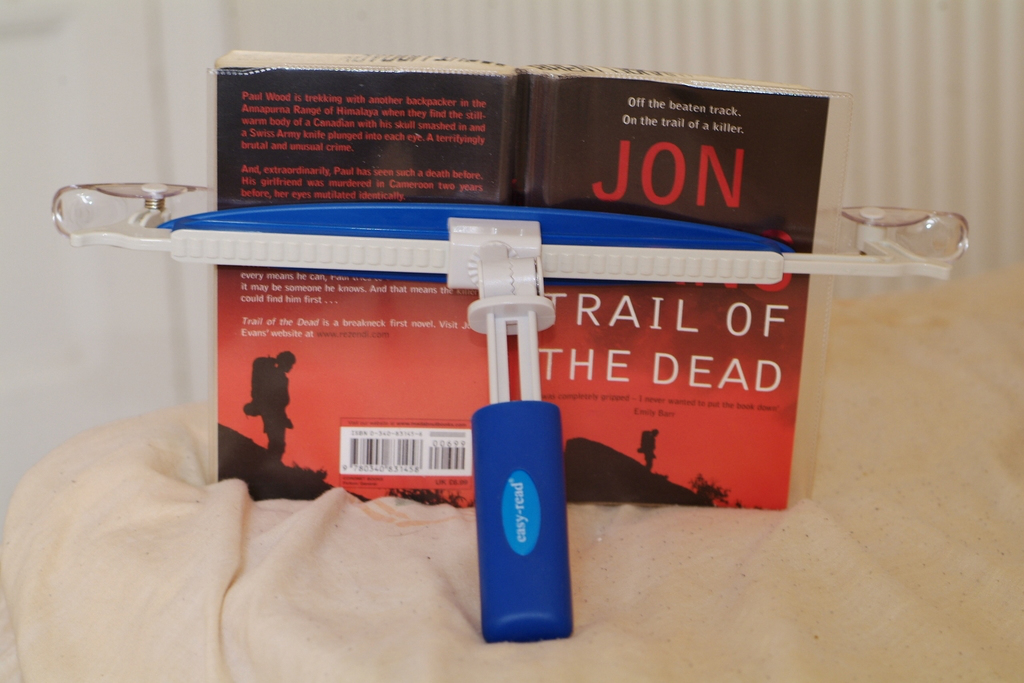}
        \caption{Chosen}
    \end{subfigure}
    \hfill
    \begin{subfigure}{0.31\linewidth}
        \centering
        \includegraphics[width=\linewidth]{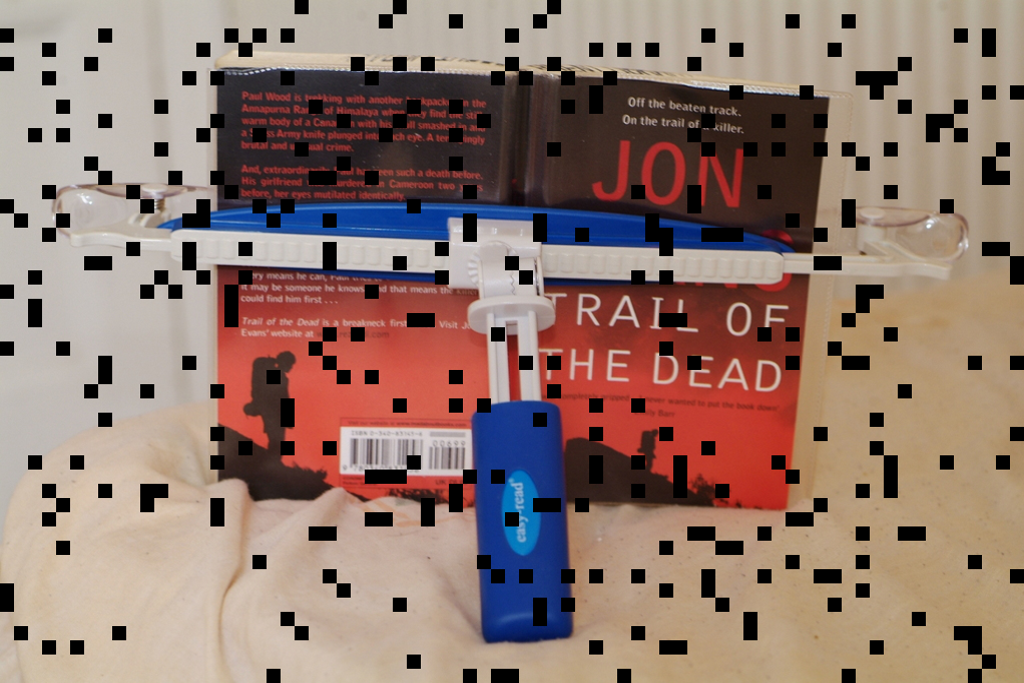}
        \caption{Blockwise}
    \end{subfigure}
    \hfill
    \begin{subfigure}{0.31\linewidth}
        \centering
        \includegraphics[width=\linewidth]{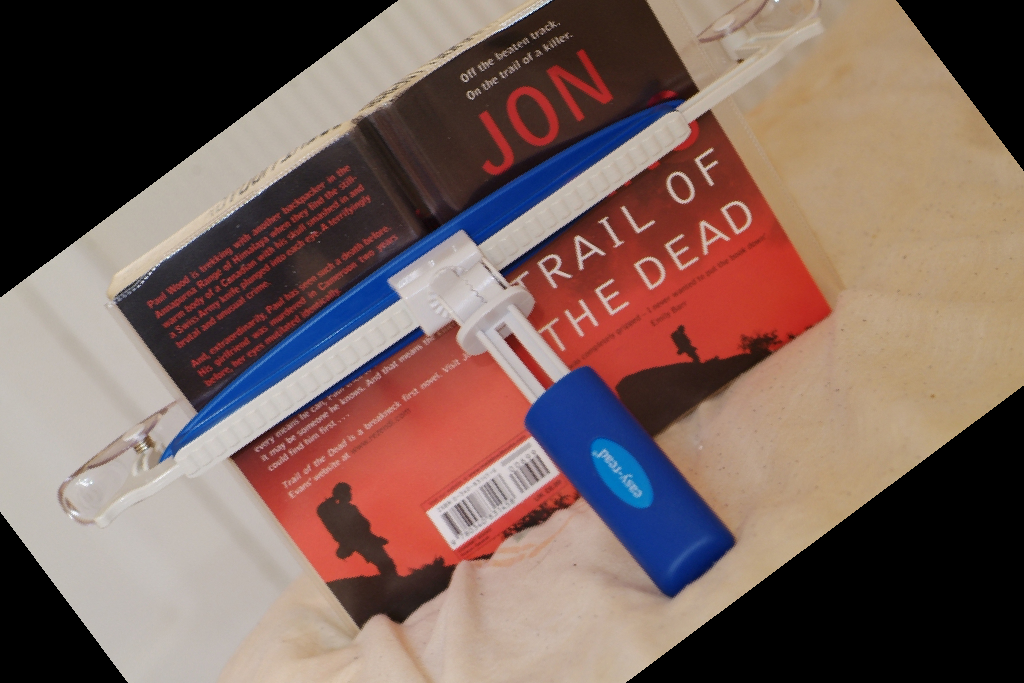}
        \caption{Rotation}
    \end{subfigure}
    
    \vspace{0.5em}

    \begin{subfigure}{0.31\linewidth}
        \centering
        \includegraphics[width=\linewidth]{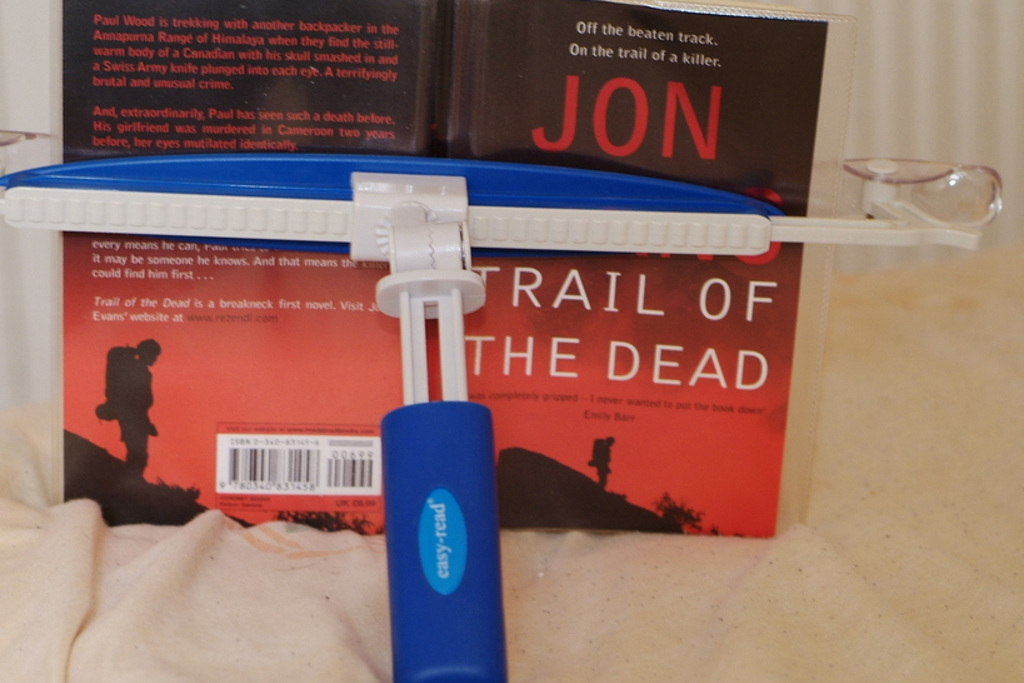}
        \caption{Crop}
    \end{subfigure}
    \hfill
    \begin{subfigure}{0.31\linewidth}
        \centering
        \includegraphics[width=\linewidth]{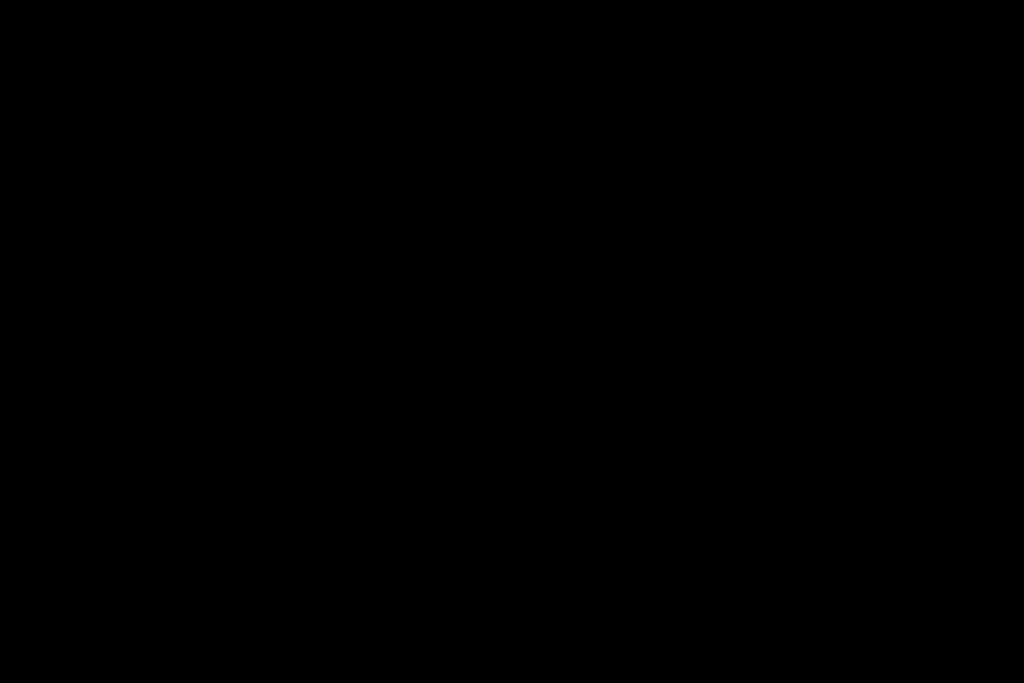}
        \caption{Black}
    \end{subfigure}
    \hfill
    \begin{subfigure}{0.31\linewidth}
        \centering
        \includegraphics[width=\linewidth]{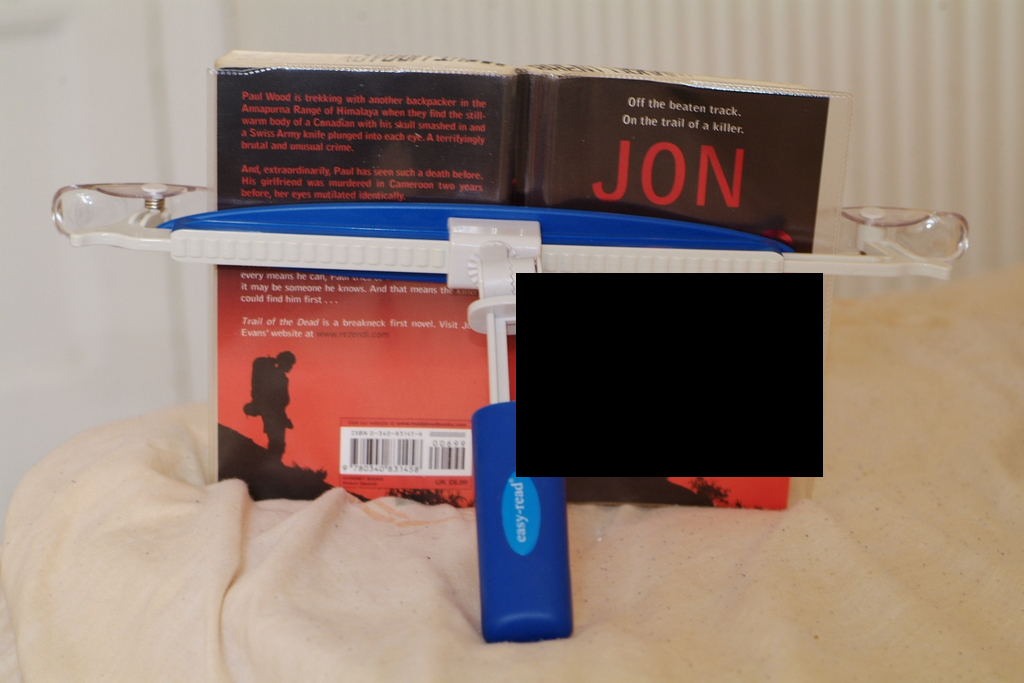}
        \caption{Ours}
    \end{subfigure}
    \vspace{-1em}
    \caption{Examples of rejected images constructed by different strategies. (a) is the chosen image.}
    \Description{Six versions of the same image compare the chosen image with block masking, rotation, cropping, complete blackening, and the proposed context-removal strategy. The proposed strategy selectively removes key visual evidence while retaining the rest of the scene.}
    \label{fig: strategies}
\end{figure}

\subsubsection{Strategies.} The existing rejected image construction strategies are listed below: (1) \textbf{Blockwise}: The chosen image is divided into blocks, with 30\% of blocks randomly masked. (2) \textbf{Rotation}: The chosen image is randomly rotated between 10 to 80 degrees. (3) \textbf{Crop}: Random cropping strategy is applied to the chosen image. (4) \textbf{Blackness}: All RGB values in the chosen image are set to 0.

\begin{table}[htbp]
  \centering
  \caption{Impact of rejected image construction strategy on mPEA-DPO. Bold indicates best.}
  % \footnotesize
  % \renewcommand\tabcolsep{2pt}
  \vspace{-1em}
  \begin{tabular}{lcccc}
    \toprule
    \multirow{2}[2]{*}{Strategy} & \multicolumn{2}{c}{MMHal Bench } & \multicolumn{2}{c}{AMBER } \\
     \cmidrule(lr){2-3}\cmidrule(lr){4-5}
          & Score$\uparrow$ & Hal.$\downarrow$ & CHAIR$\downarrow$ & HalRate$\downarrow$ \\
    \midrule
    Ours & \textbf{3.03} & \textbf{0.36} & \textbf{1.9} & \textbf{10.3} \\
    Blockwise & 2.64   & 0.42   & 2.8   & 14.1 \\
    Rotation & 2.47   & 0.46   & 2.6  & 12.5 \\
    Crop & 2.69  & 0.42   & 2.1   & 11.3 \\
    Blackness & 2.90   & 0.41   & 2.4   & 12.3 \\
    \bottomrule
  \end{tabular}%
  \label{tab: strategies}
\end{table}

\begin{figure*}[t]
    \centering
    \begin{subfigure}{0.22\linewidth} % 每个子图宽度约占一行的 1/3
        \centering
        \includegraphics[width=\linewidth]{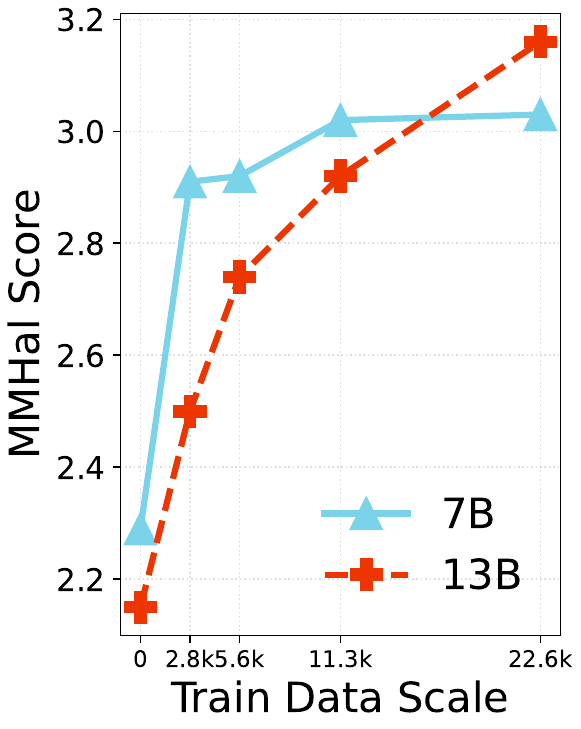}
    \end{subfigure}
    \hfill
    \begin{subfigure}{0.22\linewidth}
        \centering
        \includegraphics[width=\linewidth]{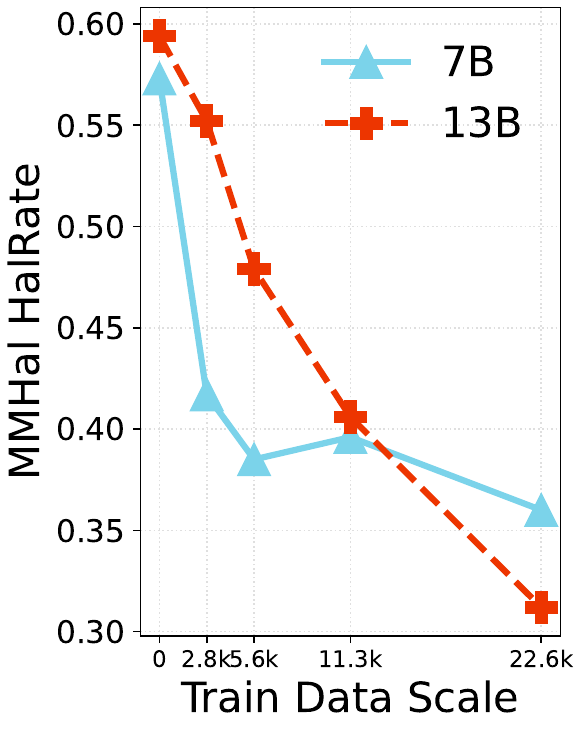}
    \end{subfigure}
    \hfill
    \begin{subfigure}{0.22\linewidth}
        \centering
        \includegraphics[width=\linewidth]{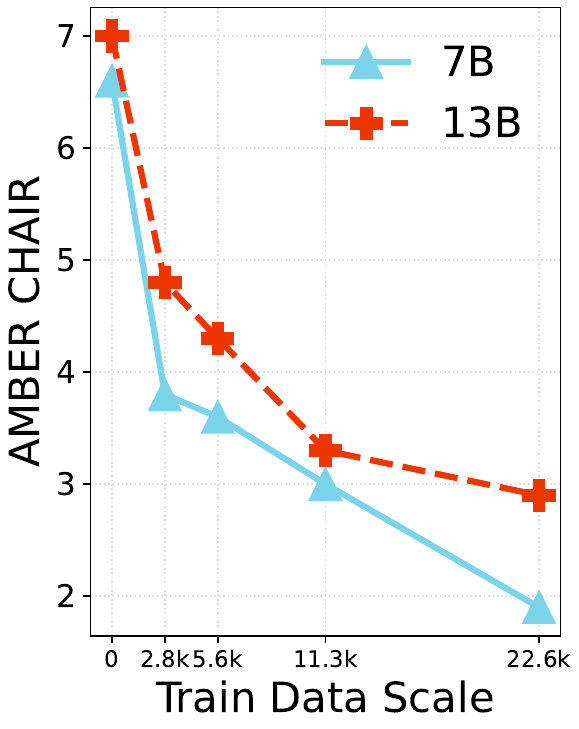}
    \end{subfigure}
    \hfill
    \begin{subfigure}{0.22\linewidth}
        \centering
        \includegraphics[width=\linewidth]{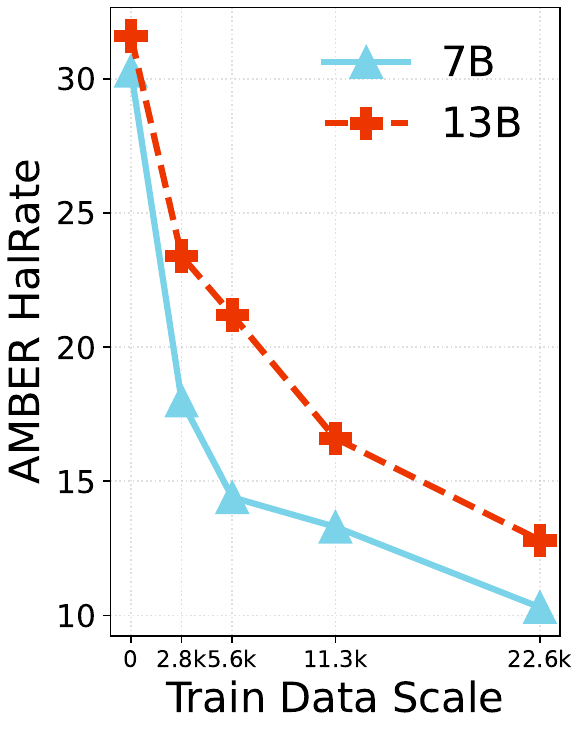}
    \end{subfigure}
    \vspace{-1em}
    \caption{\textbf{Impact of data scale} on the performance of mPEA-DPO, using LLaVA as the base model. We assess: (1) the overall score and hallucination rate on MMHalBench, and (2) CHAIR and hallucination on AMBER.}
   \label{fig:data_scale}
\end{figure*}

\subsubsection{Results.} The experimental results of mPEA-DPO under different construction strategies of rejected images are shown in Table~\ref{tab: strategies}. We observe \textbf{the rejected with removed key visual context can lead to better optimization results}. The blockwise, rotation, and crop strategies retain a significant amount of the chosen image's visual context, failing to effectively remove key visual context and thus leading to poorer performance. The blackness strategy, by completely masking the chosen image, virtually eliminates information about the chosen image, resulting in poorer performance. However, Perception-Enhanced Preference Data enhances the sensitivity of MLLMs to key visual information by removing the key visual context from the chosen images, thereby achieving the best performance.

\subsection{Impact of data scale}

To investigate the impact of data scale on mPEA-DPO, we present its performance under varying amounts of training data in Figure~\ref{fig:data_scale}. Even with only 2800 training instances, mPEA-DPO remains effective and outperforms the baseline. Additionally, we observe that the performance of mPEA-DPO consistently improves with increasing data scale, demonstrating significant performance gain.

\subsection{Attribution of Hallucination Reduction}

% \section{Attribution of reduced hallucinations}\label{coverage_reduced}

As discussed in~\citep{amirloo2024understanding}, the CHAIR metric has a known limitation in that it does not penalize shorter responses, which may trivially reduce hallucination scores.
To demonstrate that the reduction in CHAIR metrics is indeed driven by reduced hallucination rather than shorter or less informative outputs, we conducted a comparison on Object HalBench between our method and two of the strongest existing baselines, OPA-DPO~\citep{yang2025mitigating} and DAMA~\citep{lu2025damo}. We report three metrics: CHAIRs, CHAIRi, and Recall.

\begin{table}[htbp]
    \centering
    \caption{Comparison of hallucination and recall performance on Object HalBench. Bold denote the best performance.}
    \vspace{-1em}
    \begin{tabular}{lccc}
    \toprule
    Model & CHAIRs & CHAIRi & Recall \\
    \midrule
    OPA-DPO & 13.3 & 4.3 & 43.29 \\
    DAMA & 10.3 & 5.9 & \textbf{54.50} \\
    mPEA-DPO & \textbf{4.3} & \textbf{3.2} & 52.30 \\
    \bottomrule
    \end{tabular}
    \label{tab:attribution}
\end{table}

As shown in Table~\ref{tab:attribution}, mPEA-DPO achieves clearly superior hallucination metrics compared with both OPA-DPO and DAMA. Importantly, mPEA-DPO attains recall performance comparable to DAMA while substantially outperforming OPA-DPO, indicating that the improvement in CHAIR metrics is not simply due to shorter responses but rather reflects a genuine reduction in hallucination.

\subsection{Analysis of General Capability}

Preference optimization may negatively affect the model's generalization ability. In this section, we evaluate the general capabilities of MLLM enhanced with our mPEA-DPO on several widely used benchmarks, including MMStar~\citep{chen2024we}, AI2D~\citep{kembhavi2016diagram}, LLaVA-Bench~\citep{liu2024improved}, and MMMU~\citep{yue2024mmmu}. For LLaVA-Bench, we report the relative score judged by GPT-4o~\citep{gpt-4o:journals/corr/abs-2410-21276}. The results are presented in Table~\ref {tab:general_cap}.

\begin{table}[htbp]
\centering
\caption{ The general capability evaluation results. Values in \textbf{bold} denote the best performance.}
\small
\footnotesize
\begin{tabular}{lccccc}
\toprule
Model & MMStar & AI2D & LLaVA-Bench &MMMU(val) & MMMU(test) \\
\midrule
LLaVA-v1.5-7B & 30.3 & 49.1 & 82.3 & \textbf{32.1} & \textbf{35.3} \\ \midrule
+mPEA-DPO & \textbf{32.8} & \textbf{51.9} & \textbf{86.5} & 30.8 & 33.4 \\
\bottomrule
\end{tabular}%
\label{tab:general_cap}
\end{table}

% We observe that LLaVA+mPEA-DPO outperforms LLaVA on two out of three datasets. This indicates that the mPEA-DPO-enhanced LLaVA achieves slight improvements on the MMStar and AI2D datasets, while maintaining comparable performance on the MMMU dataset.

We observe that LLaVA+mPEA-DPO outperforms LLaVA on MMStar, AI2D, and LLaVA-Bench, while maintaining comparable performance on MMMU. The results suggest that visual preference optimization not only reduces hallucinations but also enhances the model's instruction-following capability. These results indicate that mPEA-DPO preserves and even slightly improves the general capabilities of the base model.

\section{Limitation and Conclusion}

\textbf{Limitation.} While mPEA-DPO yields notable gains in multimodal alignment and hallucination mitigation, several limitations remain. First, constructing perception-enhanced preference data incurs additional computation for generating and evaluating perturbed images. This overhead is manageable in our experiments. Second, due to limited computational capacity, we have not tested on the latest MLLMs (\eg Muffin). Third, mPEA‑DPO slightly reduces coverage metrics, likely reflecting a conservative generation strategy that avoids uncertain outputs.

\noindent\textbf{Conclusion.} In summary, our study uncovers two fundamental issues of Direct Preference Optimization (DPO) in multimodal settings: (1) \textbf{Across-Image Insensitivity} and (2) \textbf{Within-Image Insensitivity}. Through both theoretical analysis and empirical evaluation, we systematically characterize the inherent limitations of existing multimodal DPO methods in exhibiting \textbf{visual insensitivity}. To address these limitations, we propose Perception-Enhanced Alignment (PEA)-DPO, a framework for MLLM alignment that explicitly leverages visual preference signals in conjunction with standard multimodal DPO. Experiments on three widely-used benchmarks demonstrate that PEA-DPO substantially improves the performance of LLaVA-v1.5-7B and LLaVA-v1.5-13B, achieving strong results and surpassing other RLHF/RLAIF-based methods.

% \section{Conclusion}

% % \textbf{Limitation.} While mPEA-DPO yields notable gains in multimodal alignment and hallucination mitigation, several limitations remain. First, constructing perception-enhanced preference data incurs additional computation for generating and evaluating perturbed images. This overhead is manageable in our experiments. Second, due to limited computational capacity, we have not tested on the latest MLLMs (\eg Muffin). Third, mPEA‑DPO slightly reduces coverage metrics, likely reflecting a conservative generation strategy that avoids uncertain outputs.

% In summary, our study uncovers two fundamental issues of Direct Preference Optimization (DPO) in multimodal settings: (1) \textbf{Across-Image Insensitivity} and (2) \textbf{Within-Image Insensitivity}. Through both theoretical analysis and empirical evaluation, we systematically characterize the inherent limitations of existing multimodal DPO methods in exhibiting \textbf{visual insensitivity}. To address these limitations, we propose Perception-Enhanced Alignment (PEA)-DPO, a framework for MLLM alignment that explicitly leverages visual preference signals in conjunction with standard multimodal DPO. Experiments on three widely-used benchmarks demonstrate that PEA-DPO substantially improves the performance of LLaVA-v1.5-7B and LLaVA-v1.5-13B, achieving strong results and surpassing other RLHF/RLAIF-based methods.

% \input{rebuttal/rebuttal}

\balance
\bibliographystyle{ACM-Reference-Format}
\bibliography{reference}


% \input{section/appendix}

\end{document}